%% file: _main.tex
\input{_constants}
\arxiv 

\pdfoutput=1
\documentclass[10pt,twocolumn,letterpaper]{article}
\input{cvpr_header}
\begin{document}

\title{\paperTitle}
\author{\authorBlock}

\twocolumn[{
\renewcommand\twocolumn[1][]{#1}
\maketitle
\begin{center}
    \centering
    \vspace*{-.8cm}
    \includegraphics[width=\textwidth]{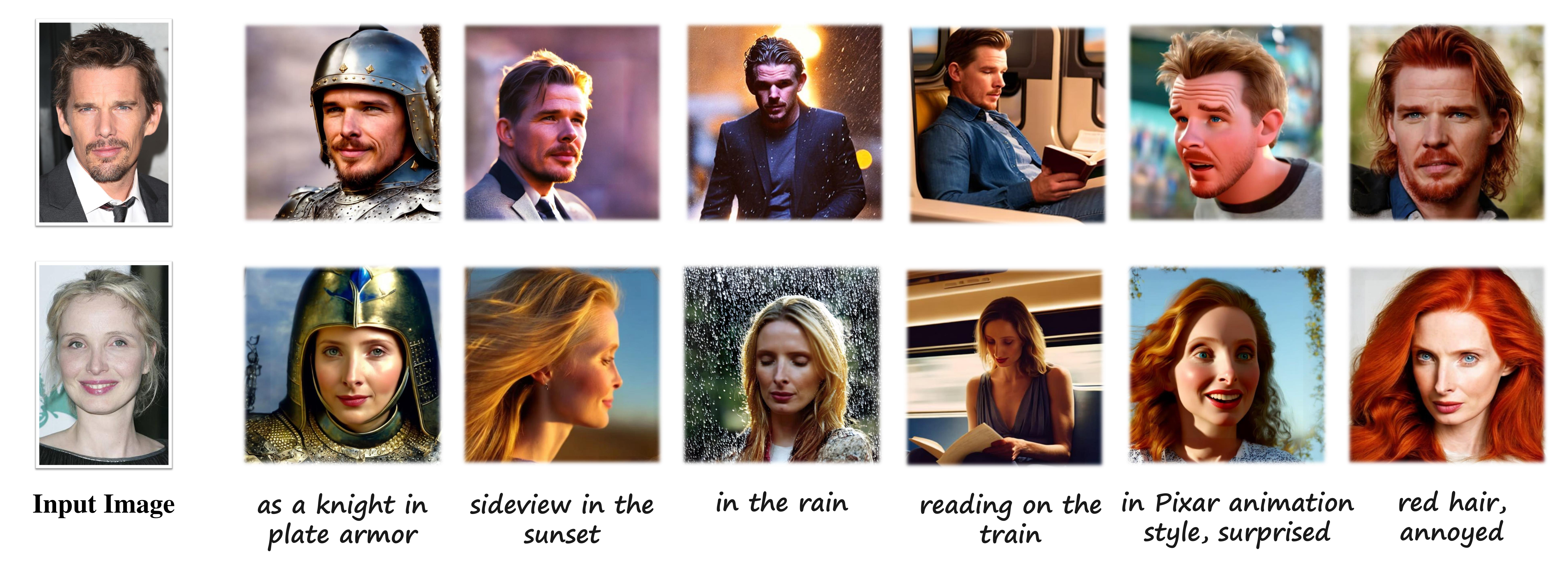}
    \vspace*{-.6cm}
    \captionof{figure}{Given a single facial photo of as the reference and a text prompt, our proposed method can generate images in a variety of styles, angles, and expressions without any test-time fine-tuning at the inference stage. The results exhibit dressing-up modifications, viewpoint control, recontextualization, art renditions, property alteration, as well as emotion integration, while preserving high fidelity to the face.}
\label{fig::overview}
\end{center}
}]

\input{00_abstract}
\input{01_intro}
\input{02_related}

\input{03_method}
\input{04_experiment}
\input{10_conclusion}

{\small
\bibliographystyle{ieeenat_fullname}
\bibliography{11_references}

}

\ifarxiv \clearpage \appendix \input{12_appendix} \fi

\end{document}

%% file: _constants.tex
\def\paperID{12309}
\def\confName{CVPR}
\def\confYear{2024}

\def\paperTitle{IDAdapter: Learning Mixed Features for Tuning-Free Personalization of Text-to-Image Models}

\def\authorBlock{
    Siying Cui\textsuperscript{1,3} \quad
    Jia Guo\textsuperscript{2}\textsuperscript{*}\quad
    Xiang An\textsuperscript{2,3}\quad
    Jiankang Deng\textsuperscript{2} \quad
    Yongle Zhao\textsuperscript{3}\quad
    Xinyu Wei\textsuperscript{1}\quad
    Ziyong Feng\textsuperscript{3}
     \\
    \textsuperscript{1}Peking University\qquad
    \textsuperscript{2}InsightFace\qquad
    \textsuperscript{3}DeepGlint\qquad
    \\
    {\tt\small {\textsuperscript{1}hycsy@stu.pku.edu.cn\qquad \textsuperscript{*}Corresponding Author: guojia@gmail.com
    }}
}

\newif\ifreview 
\newif\ifarxiv \newcommand{\arxiv}{\arxivtrue}
\newif\ifcamera 
\newif\ifrebuttal 

%% file: cvpr_header.tex
\ifreview \usepackage[review]{cvpr} \fi
\ifarxiv \usepackage[pagenumbers]{cvpr} \fi
\ifrebuttal \usepackage[rebuttal]{cvpr} \fi
\ifcamera \usepackage{cvpr} \fi

\input{_macros}  

\usepackage{xr-hyper}

\makeatletter
\newcommand*{\addFileDependency}[1]{
  \typeout{(#1)}
  \@addtofilelist{#1}
  \IfFileExists{#1}{}{\typeout{No file #1.}}
}

\makeatother

\definecolor{cvprblue}{rgb}{0.21,0.49,0.74}
\usepackage[pagebackref,breaklinks,colorlinks,citecolor=cvprblue]{hyperref}
\usepackage{indentfirst}
\usepackage[capitalize]{cleveref}
\crefname{section}{Sec.}{Secs.}
\crefname{table}{Table}{Tables}
\crefname{figure}{Fig.}{Figs.}

%% file: _macros.tex
\usepackage{graphicx}	
\usepackage{amsmath}	
\usepackage{amssymb}	
\usepackage{booktabs}
\usepackage{times}
\usepackage{microtype}
\usepackage{epsfig}
\usepackage[table,xcdraw,dvipsnames]{xcolor}
\usepackage{caption}
\usepackage{float}
\usepackage{placeins}
\usepackage{color, colortbl}
\usepackage{stfloats}
\usepackage{enumitem}
\usepackage{tabularx}
\usepackage{xstring}
\usepackage{multirow}
\usepackage{xspace}
\usepackage{url}
\usepackage{subcaption}
\usepackage{xcolor}
\usepackage[hang,flushmargin]{footmisc}
\usepackage{makecell}
\ifcamera \usepackage[accsupp]{axessibility} \fi

\ifarxiv  \fi

\newcommand{\R}[1]{{%
    \textbf{%
        \ifstrequal{#1}{1}{\textcolor{red}{R#1}}{%
        \ifstrequal{#1}{2}{\textcolor{blue}{R#1}}{%
        \ifstrequal{#1}{3}{\textcolor{magenta}{R#1}}{%
        \ifstrequal{#1}{4}{\textcolor{teal}{R#1}}{%
                           \textcolor{cyan}{R#1}%
        }}}}%
    }%
}}

%% file: 00_abstract.tex
\begin{abstract}

\vspace{-.2in}

Leveraging Stable Diffusion for the generation of personalized portraits has emerged as a powerful and noteworthy tool, enabling users to create high-fidelity, custom character avatars based on their specific prompts. However, existing personalization methods face challenges, including test-time fine-tuning, the requirement of multiple input images, low preservation of identity, and limited diversity in generated outcomes. To overcome these challenges, we introduce IDAdapter, a tuning-free approach that enhances the diversity and identity preservation in personalized image generation from a single face image. IDAdapter integrates a personalized concept into the generation process through a combination of textual and visual injections and a face identity loss. During the training phase, we incorporate mixed features from multiple reference images of a specific identity to enrich identity-related content details, guiding the model to generate images with more diverse styles, expressions, and angles compared to previous works. Extensive evaluations demonstrate the effectiveness of our method, achieving both diversity and identity fidelity in generated images.

\end{abstract}

%% file: 01_intro.tex
\section{Introduction}
\label{sec:intro}

Recently, the field of text-to-image (T2I) synthesis has witnessed significant advancements, especially with the advent of diffusion models. Models such as Imagen proposed by~\cite{saharia2022photorealistic}, DALL-E2 by ~\cite{ramesh2022hierarchical}, and Stable Diffusion by ~\cite{sd} have gained attention for their ability to generate realistic images from natural language prompts. While these models excel in generating complex, high-fidelity images from extensive text-image datasets, the task of generating images of specific subjects from user-provided photos remains a significant challenge.

Personalization in text-to-image (T2I) synthesis has been primarily achieved through methodologies employing pre-trained models, as outlined in works such as \cite{gal2022image,ruiz2022dreambooth,kumari2023multi,tewel2023key,avrahami2023break,hao2023vico,smith2023continual}. These methods typically involve fine-tuning the models with a set of specific reference images (ranging from 3 to 20). While effective, this approach calls for specialized training of certain network layers, often requiring considerable computational resources and extended processing times on advanced GPUs, which may not be feasible for user-centric applications. An alternative strategy, discussed in studies like~\cite{xiao2023fastcomposer,wei2023elite,chen2023subject,chen2022re}, involves augmenting pre-trained diffusion models with additional parameters like adapters trained on large personalized image datasets. This approach enables tuning-free conditional generation but typically lacks the fidelity and diversity of fine-tuning methods. For example, as indicated in \cite{chen2023photoverse} and \cite{shi2023instantbooth}, this approach often restricts the generated images to the expressions present in the input image, thus limiting the expansive creative potential of diffusion models.

Drawing inspiration from test-time fine-tuning methods utilizing multiple reference images and the adapter series as described in works~\cite{mou2023t2i,ye2023ip,wang2023styleadapter}, we introduce IDAdapter. This innovative approach synthesizes features from various images of a person during training, effectively mitigating overfitting to non-identity attributes. IDAdapter operates by freezing the base diffusion model's primary weights, with under 10 hours of training on a single GPU. During inference, IDAdapter requires only a single reference image and textual prompts to produce diverse, high-fidelity images that maintaining the person's identity, as depicted in Figure \ref{fig::overview}. It broadens the range of what the base model can generate, making the results more diverse while preserving identity, which surpasses the limitations of previous models. Our contributions are threefold:
\begin{enumerate}
\item We present a method that incorporates mixed features from multiple reference images of the same person during training, yielding a T2I model that avoids the need for test-time fine-tuning.
\item This technique, without test-time fine-tuning, can generate varied angles and expressions in multiple styles guided by a single photo and text prompt, a capability not previously attainable.
\item Comprehensive experiments confirm that our model outperforms earlier models in producing images that closely resemble the input face, exhibit a variety of angles, and showcase a broader range of expressions.
\end{enumerate}

%% file: 02_related.tex
\section{Related Work}
\label{sec:related}

\subsection{Text-to-Image Models}

The field of computational image generation has witnessed remarkable advancements due to the evolution of deep generative models for text-to-image synthesis. Techniques like Generative Adversarial Networks (GANs) \cite{xia2021tedigan, kang2023scaling}, auto-regressive models \cite{ramesh2021zero}, and diffusion models \cite{ho2020denoising, sd} have played a crucial role. Initially, these models were limited to generating images under specific domains and textual conditions. However, the introduction of large-scale image-text datasets and advanced language model encoders has significantly improved text-to-image synthesis capabilities. The pioneering DALL-E~\cite{ramesh2021zero} utilized autoregressive models for creating diverse images from text prompts. This was followed by GLIDE~\cite{nichol2021improved}, which introduced more realistic and high-resolution images using diffusion models. Consequently, diffusion models have increasingly become the mainstream method for text-to-image synthesis. Recent developments like DALL-E 2~\cite{ramesh2022hierarchical}, Imagen~\cite{saharia2022photorealistic}, and LDM~\cite{sd} have further enhanced these models, offering more realism, better language understanding and diverse outputs. The success of Stable Diffusion~\cite{sd} in the open-source community has led to its widespread use and the development of various fine-tuned models. Our methodology, acknowledging this trend, is based on the Stable Diffusion model.

\subsection{Personalization via Subject-Driven Tuning}
The goal of personalized generation is to create variations of a specific subject in diverse scenes and styles based on reference images. Originally, Generative Adversarial Networks (GANs) were employed for this purpose, as illustrated by~\cite{nitzan2022mystyle}, who achieved personalization by fine-tuning StyleGAN with around 100 facial images. Subsequently, pivotal tuning \cite{roich2022pivotal}, which involved fine-tuning latent space codes in StyleGAN, enabled the creation of variant images. However, these GAN-based methods faced limitations in subject fidelity and style diversity. Recent advancements have been made with the Stable Diffusion Model, offering improvements in subject fidelity and outcome diversity. Textual Inversion~\cite{gal2022image} optimized input text embeddings with a small set of images for subject image generation. The study by~\cite{voynov2023p+} enhanced textual inversion to capture detailed subject information. DreamBooth \cite{ruiz2022dreambooth} optimized the entire T2I network for higher fidelity. Following this, several methods like CustomDiffusion~\cite{kumari2023multi}, SVDiff~\cite{han2023svdiff}, LoRa~\cite{db_lora,hu2021lora}, StyleDrop~\cite{sohn2023styledrop}, and the approach by ~\cite{houlsby2019parameterefficient} proposed partial optimizations. DreamArtist \cite{dong2022dreamartist} demonstrated style personalization with a single image. Despite their effectiveness, these methods involve time-consuming multi-step fine-tuning for each new concept, limiting their practicality in real-world applications.

\subsection{Tuning-Free Text-to-Image Personalization}
A distinct research direction involves training models with extensive domain-specific data, thereby eliminating the need for additional fine-tuning at the inference stage. Facilitating object replacement and style variation, InstructPix2Pix~\cite{brooks2023instructpix2pix} integrates latent features of reference images into the noise injection process. ELITE~\cite{wei2023elite} introduced a training protocol combining global and local mappings, utilizing the OpenImages test set. UMM-Diffusion~\cite{ma2023unified}, leveraging LAION-400M dataset~\cite{schuhmann2021laion}, proposed a multimodal latent diffusion approach that combines text and image inputs. Several studies, such as UMM~\cite{ma2023unified}, ELITE~\cite{wei2023elite}, and SuTI~\cite{chen2023subject}, have demonstrated subject image generation without fine-tuning. Similarly, Taming-Encoder~\cite{jia2023taming} and InstantBooth~\cite{shi2023instantbooth} focus on human and animal subjects, employing a new conditional branch for diffusion models. FastComposer~\cite{xiao2023fastcomposer}, Face0~\cite{valevski2023face0} and PhotoVerse~\cite{chen2023photoverse} have also contributed novel approaches in this domain. Despite these advancements, a key challenge remains in balancing ease of use with generation quality and diversity. Our proposed solution, IDAdapter, addresses this issue by coordinating model usability with output quality and diversity.

%% file: 03_method.tex
\section{Method}
\label{sec:method}
\begin{figure*}[tp]
    \centering
    \includegraphics[width=\linewidth]{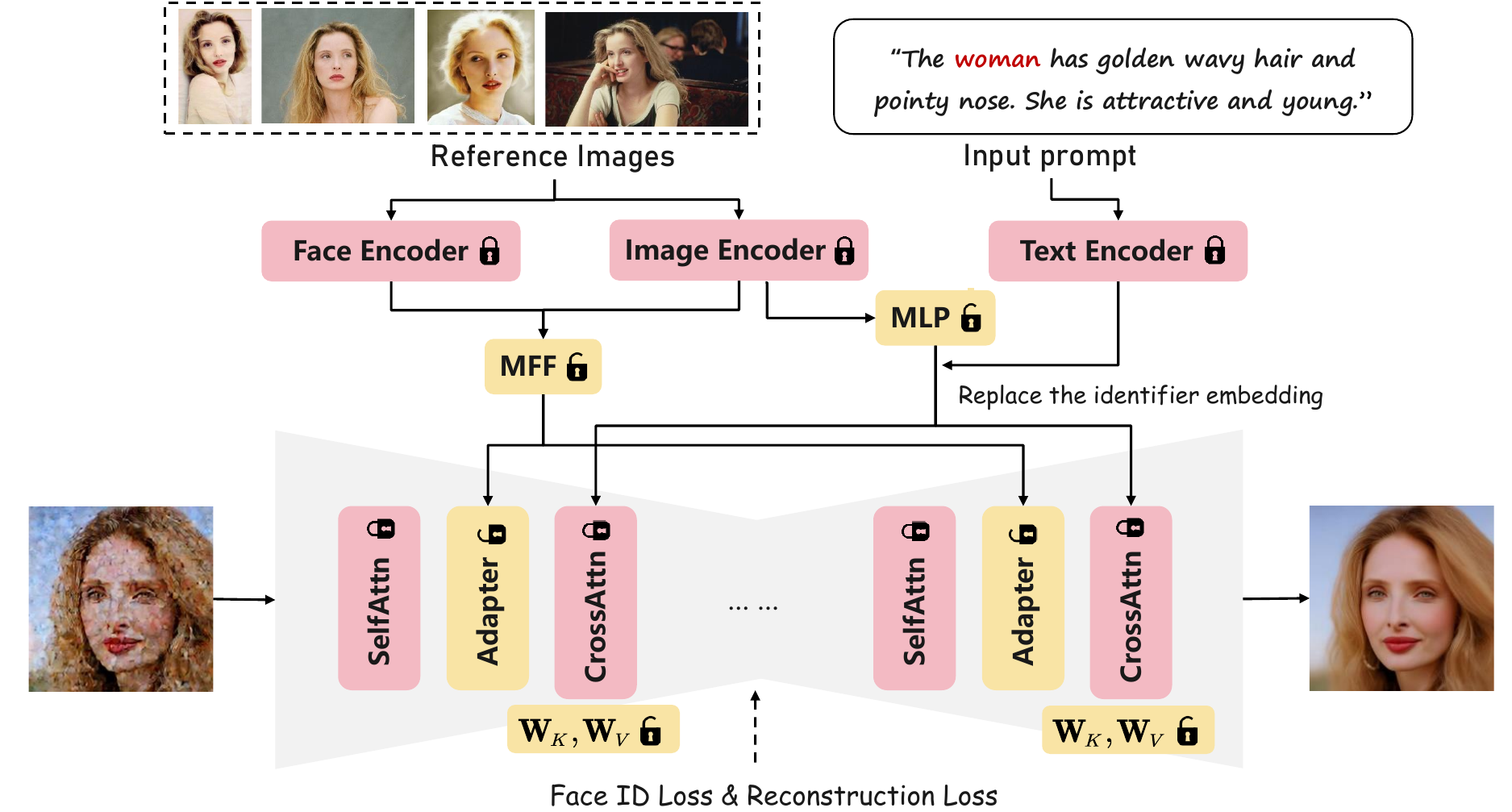}
    \caption{\textbf{The overview of IDAdapter training}. In each optimization step, we randomly select $N$ different images of the same identity.  We label the faces of all the reference images ``\textit{[class noun]}" (e.g. ``\textit{woman}", ``\textit{man}", etc.), and regard the text description and the reference images as a training pair. The features extracted from the reference images are then fused using a mixed facial features (MFF) module, which provides the model with rich detailed identity information and possibilities for variation. At the inference stage, only a single image is required, which is replicated to form a set of $N$ reference images.}
    \label{fig::pipeline}
\end{figure*}

Given only a single face image of a specific person, we intend to generate a range of vivid images of the person guided by text prompts with diversity. Example diversity includes not only adjusting dressing-up, properties, contexts, and other semantic modifications (these attributes are referred to as "styles" in this paper), but generating various facial expressions and poses. We next briefly review the necessary notations of Latent Diffusion Models (Sec. \ref{sec:pre}) as well as the method for simply extracting facial features from a single image (Sec. \ref{sec:iie}), then present our technique to extract mixed facial features from a few images (Sec. \ref{sec::mff}), and finally incorporate it as a new concept to the generator structure by the adapter layers (Sec. \ref{sec:integ}). Fig. \ref{fig::pipeline} shows the overview of our approach based on the Stable Diffusion \cite{sd} structure. 

\subsection{Preliminaries}
\label{sec:pre}

Text-to-Image (T2I) diffusion models $\mathbf\epsilon _\theta$ denoise a noise map $\mathbf\epsilon \in \mathbb{R}^{h \times w}$ into an image $\mathbf x_0$ based on a textual prompt $\mathit T$. In this work, we utilize Stable Diffusion, a specific instance of Latent Diffusion Model (LDM), which comprises three key components: an image encoder, a decoder, and an iterative UNet denoising network for processing noisy latent representations.

The encoder $\mathcal E$ maps an image $\mathbf x_0$ from the pixel space to a low-dimensional latent space $\mathbf z=\mathcal E(\mathbf x_0)$, while the decoder $\mathcal D$ reconstructs the latent representation $\mathbf z$ back into an image to achieve $\mathcal D(\mathcal E (\mathbf x_0))\approx \mathbf x_0$. The diffusion model incorporates an input text embedding $\mathbf{C}=\Theta(T)$, which is generated using a text encoder $\Theta$ and a text prompt $\mathit T$ and then employed in the intermediate layers of the UNet through a cross-attention mechanism:
\begin{equation}
\operatorname{Attention}(\mathbf Q, \mathbf K, \mathbf V)=\operatorname{softmax}\left(\frac{\mathbf Q \mathbf K^T}{\sqrt{d}}\right) \cdot \mathbf V
\end{equation}
where $\mathbf Q=\mathbf W_Q \cdot \varphi\left(\mathbf z_t\right)$, $\mathbf K=\mathbf W_K \cdot \mathbf{C}$, $\mathbf V=\mathbf W_V \cdot \mathbf{C}$, $ \varphi\left(\mathbf z_t\right)$ is the hidden states through the UNet implementation, $d$ is the scale factor utilized for attention mechanisms. The training goal for the latent diffusion model is to predict the noise added to the image's latent space, a formulation denoted as:
\begin{equation}
\mathcal{L}_{\mathrm{SD}}=\mathbb{E}_{\mathbf z\sim\mathcal{E}\left(\mathbf x_0\right), \mathbf{C}, \mathbf \epsilon \sim \mathcal{N}(0,1), t}\left[\left\|\mathbf \epsilon-\epsilon_\theta\left(\mathbf z_t, t, \mathbf{C}\right)\right\|_2^2\right]
\end{equation}
where  $\mathbf \epsilon$ is the ground-truth noise, $\mathbf z_t$ is noisy latent representations at the diffusion process timestep $t$. See  \cite{sd} for more details.

\subsection{Facial Features}
\label{sec:iie}
Our objective is to extract facial features from input images, inject them with the stylistic information denoted by text prompts, and generate a rich variety of images with fidelity to the identified facial characteristics. Intuitively, this diversity includes at least the following three aspects: A) Diversity of styles, where the generated images must conform to the styles indicated by the prompts; B) Diversity in facial angles, signifying the capability to produce images of the person from various facial poses; C) Diversity of expressions, which refers to the ability to generate images of the person displaying a range of different expressions or emotions.

An intuitive approach is learning the features of input facial images in the textual space and embedding these features into the generative guiding process of Stable Diffusion, so that we can control the image generation of the person via a specific identifier word. However, as noted by several studies  \cite{chen2023photoverse,hyung2023magicapture,ma2023subjectdiffusionopen,shi2023instantbooth}, the sole use of textual space embeddings constrains the ultimate quality of generated images. A potential cause for this pitfall could be the limitation of the textual space features in capturing identity (ID) characteristics. Consequently, it becomes imperative to supplement textual conditional guidance with guidance based on image features to augment the image generation capabilities.

\begin{figure}[tp]
    \centering
    \includegraphics[width=\linewidth]{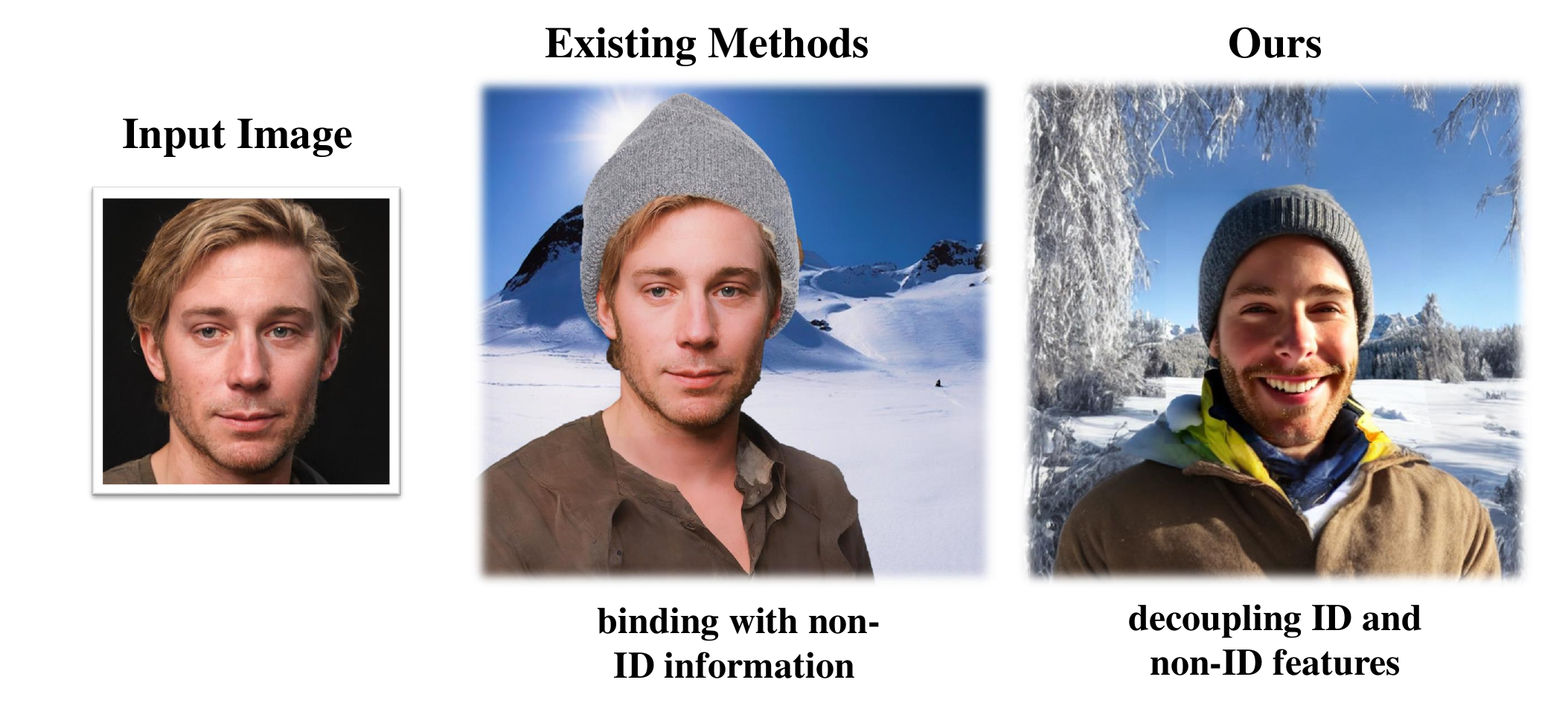}
    \caption{\textbf{Binding non-identity (non-ID) information vs. decoupling ID and non-ID information.} Most of the existing generation methods bind the identifier word to non-ID information and rarely exhibit changes in facial expressions, lighting, poses, etc. Our method decouples ID and non-ID information and can generate high-fidelity images with diversity of styles, expressions, and angles (text prompt of the example: ``\textit{man in the snow, happy}")}
    \label{fig::binding}
\end{figure}

We find that both commonly used general CLIP image encoders and feature vector encoders from face recognition networks exhibit a strong binding with non-identity (non-ID) information of the input images, such as facial poses and expressions. This binding results in the generated images lacking diversity at the person level, as illustrated in Figure \ref{fig::binding}. To address this issue, we propose the Mixed Facial Features module (MFF). This module is designed to control the decoupling of ID and non-ID features during the generation process of the diffusion model, thereby enabling the generation of images with enhanced diversity.

\subsection{Mixed Facial Features (MFF)}
\label{sec::mff}
\begin{figure}[tp]
    \centering
    \includegraphics[width=\linewidth]{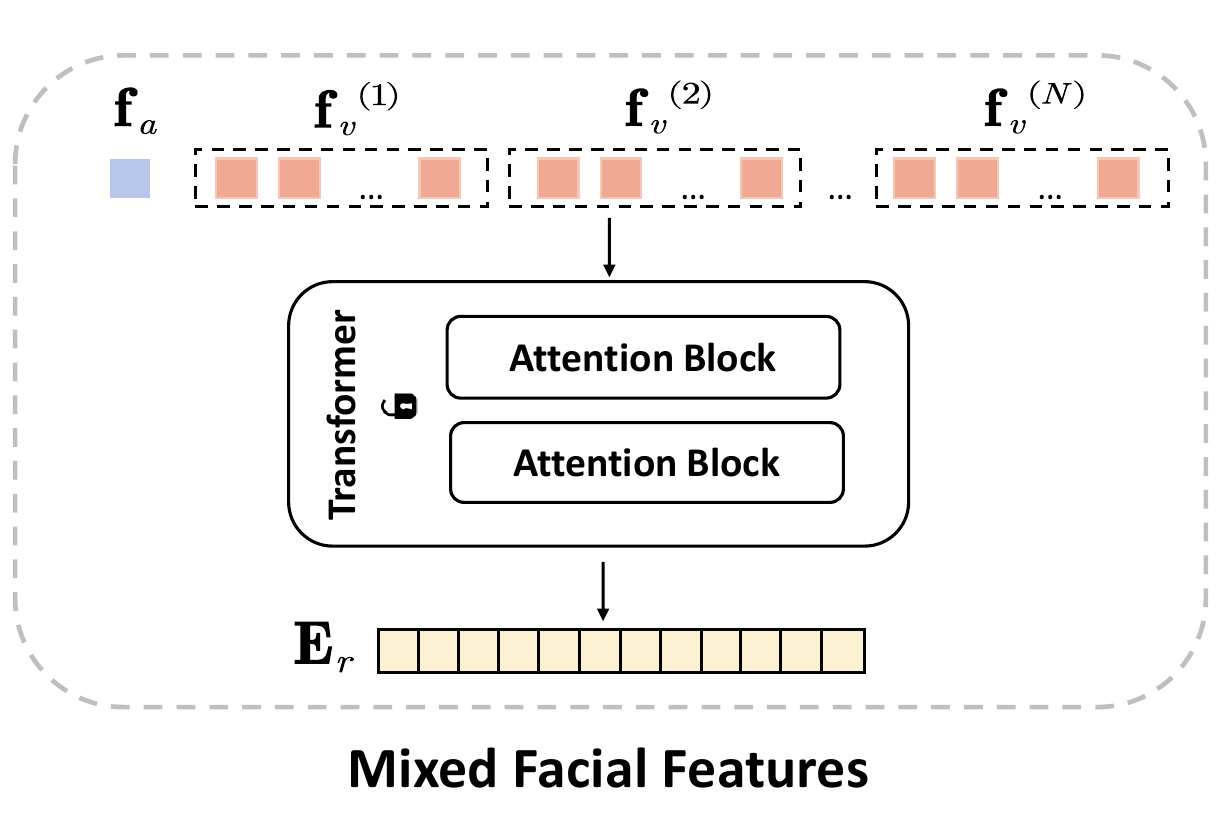}
    \caption{\textbf{Architecture of MFF:} Our MFF consists of a learnable transformer implemented with two attention blocks that translates identity feature $\textbf f_a$ and patch feature $\textbf f_v$ into a latent MFF vision embedding $\textbf E_r$, which will be injected to the self-attention layers of the UNet through adapters.}
    \label{fig::mff}
\end{figure}

The core idea behind \textbf{MFF} is to utilize rich detailed information from multiple reference images to help IDAdapter better extract identity features and achieve face fidelity, rather than simply learn from a single face paste. Specifically, we combine the features of $N$ face images  $\{\mathbf x^{(1)}, \mathbf x^{(2)},...,\mathbf x^{(N)}\}$ with the prompt $T$ to guide the generation of Stable Diffusion, where $\mathbf x^{(i)}\in \mathbb{R}^{h\times w\times c}$ for $i=1,..,N$, $(h,w)$ is the resolution of $\mathbf x^{(i)}$ and $c$ is the number of channels. We illustrate the idea in Figure \ref{fig::mff}.

Given a reference image set containing $K$ images, we first enrich them to $N$ images $\{\mathbf x^{(1)}, \mathbf x^{(2)},..,\mathbf x^{(K)},..,\mathbf x^{(N)}\}$ if $K<N$, through various data augmentation operations such as random flipping, rotating, and color transformations. We first encode all of the reference images into visual patch features $\{\mathbf f_v^{(1)},\mathbf f_v^{(2)},..,\mathbf f_v^{(N)}\}$ and a learned class embedding $\{\mathbf f_{c}^{(1)},\mathbf f_{c}^{(2)},..,\mathbf f_{c}^{(N)}\}$ using the vision model of CLIP \cite{radford2021learning}, where  $\mathbf f_v^{(i)} \in \mathbb{R}^{p^2 \cdot c\times d_v}$  and $\mathbf f^{(i)}_{c}\in  \mathbb{R}^{1\times d_v}$ . Here, $(p,p)$ is the patch size, and $d_v$ is the dimension of these embedded patches through a linear projection. Then we obtain an enriched patch feature $\mathbf f_v$ by concatenating the patch features from all the reference images. We have:
\begin{equation}
\mathbf f_v=Concat(\{\mathbf f_v^{(i)}\}^N_1)
\end{equation}

This enriched feature $\mathbf f_v$ is derived from multiple images under the same identity, so their common characteristics (i.e., the identity information) will be greatly enhanced, while others (such as the face angle and expression of any specific image) will be somewhat weakened. Therefore, $\mathbf f_v$ can greatly assist in diversifying the generation results as indicated in Sec. \ref{sec::ablation}.
We find that with $N=4$, personalization results are strong and maintain identity fidelity, editability and facial variation. 

To further guarantee the identity, we encode the faces from all the enriched reference images  $\{\mathbf x^{(1)}, \mathbf x^{(2)},..,\mathbf x^{(N)}\}$ into identity features  $\{\mathbf f_a^{(1)},\mathbf f_a^{(2)},..,\mathbf f_a^{(N)}\}$ using the face encoder of Arcface face recognition approach \cite{deng2018arcface}, where $\mathbf f_a^{(i)} \in \mathbb{R}^{1\times d_{a}}$ for $i=1,..,N$. Then we calculate the average feature vector $\mathbf f_a$ as an identity feature. We have:
\begin{equation}
\label{eq:fa}
\mathbf f_a=\sum_{i=1}^N \mathbf f_a^{(i)}/ N
\end{equation}

Then we appends the identity feature $\mathbf f_a$ to the patch feature $\mathbf f_v$ as one and embed it into a learnable lightweight transformer $\mathcal P_{\text{visual}}$ implemented with two attention blocks as illustrated in Figure \ref{fig::mff}. We have:

\begin{equation}
\mathbf E_{r}=\mathcal P_{\text{visual}}([\mathbf f_v,\mathbf f_a])
\end{equation}

Finally we obtain a MFF vision embedding $\mathbf E_r$, which compresses the facial information and is adapted to the latent space of the diffusion model. The feature $\mathbf E_r$  will be injected into the self-attention layers of the UNet through adapters.

\subsection{Personalized Concept Integration}
\label{sec:integ}

\noindent\textbf{Textual Injection} In addition to obtaining mixed facial features from the pixel space, we also aim at injecting a new personalized concept into Stable Diffusion's ``dictionary". We label the faces of all the reference images ``\textit{[class noun]}" (e.g. ``\textit{woman}", ``\textit{man}", etc.), which can be specified by the user, and denote ``\textit{sks}" as an identifier word. In this paper, we assume that ``\textit{[class noun] is sks}" is appended to the end of each prompt by default, thereby linking the face features with the identifier word. As mentioned in the approach to generate patch features in Sec. \ref{sec::mff} using the vision model of CLIP \cite{radford2021learning}, we also obtain a learned class embedding $\{\mathbf f_{c}^{(1)},\mathbf f_{c}^{(2)},..,\mathbf f_{c}^{(N)}\}$ simultaneously. We adopt their average embedding to map all the reference images to a compact textual concept through a learnable multi-layer perceptron $\mathcal P_{\text{textual}}$:
\begin{equation}
\mathbf E_{c}=\mathcal P_{\text{textual}}(\sum_{i=1}^N \mathbf f_c^{(i)}/ N)
\end{equation}
where $\mathbf E_c$ is the identity text embedding of the reference images, projected from the visual space to the textual space of Stable Diffusion in essence. At the first embedding layer of the text encoder, we replace the text embedding of the identifier word ``\textit{sks}" with the identity text embedding $\mathbf E_c$ to inject textual personalized concept into the UNet. This final text embedding will be the condition in the cross-attention layers of Stable Diffusion.

\vspace{\baselineskip}\noindent\textbf{Visual Injection}\quad We find that the model tends to generate overfitting results (e.g. fixed expressions, poses) if we fine-tune the entire Stable Diffusion since the prior is ruined. This motivates the need for key parameters to learn the personalized concept with the output of MFF. In this regard, some existing research \cite{gal2023encoder,kumari2023multi} have emphasized the significance of attention layers. Therefore, our approach is to extend the model with trainable adapter layers and optimize query and key matrices $\mathbf W_K$, $\mathbf W_V$ in the cross-attention modules. 

Specifically, as for the injection of the MFF vision embedding $\mathbf E_{r}$, we employ a new learnable adapter layer between each frozen self-attention layer and cross-attention layer: 
\begin{equation}
\mathbf y:=\mathbf y+\beta \cdot \tanh (\gamma) \cdot S\left(\left[\mathbf y, \mathbf E_{r}\right]\right)
\end{equation}
where, $\mathbf y$ is the output of the self-attention layer, $S$ is the self-attention operator, $\gamma$ is a learnable scalar initialized as 0, and $\beta$ is a constant to balance the importance of the adapter layer.

Then, by updating the key and value projection matrices in each cross-attention block, the model is able to focus on the visual characteristics of the face and link them with the personalized concept in the textual space.

\vspace{\baselineskip}\noindent\textbf{Face Identity Loss}\quad Our experiments will show the diversity of generation achieved by learning mixed face features, which looses the regularization of facial region. However, it gives rise to the problem of identity preservation. Accordingly, we introduce a face identity loss $\mathcal{L}_\mathrm{id}$ that supervises the model to preserve the identity of reference images. This allows the model to generate diverse appearances, as well as retain the identity feature. Specifically, we utilize a pretrained face recognition model $\mathcal{R}$ \cite{deng2018arcface} :
\begin{equation}
\label{eq:lid}
\mathcal{L}_\mathrm{i d}=\mathbb{E}_{\mathbf{\hat{x}}_0}\left[1-\cos \left(\mathcal{R}\left(\mathbf{\hat{x}}_0\right),\mathbf f_a\right)]\right.
\end{equation}
where $\cos$ denotes the cosine similarity, $\mathbf{\hat{x}}_0$ is the predicted denoised image sample based on the model output $\mathbf z_t$ at the diffusion timestep $t$, and $\mathbf f_a$ is  the average identity feature calculated by Equation \ref{eq:fa}. To prevent an unclear face of image $\mathbf{\hat{x}}_0$ misleading the model, we utilize a face detection model \cite{deng2019retinaface} $\mathcal F$. Face identity loss is applied only when a face is detected  in  $\mathbf{\hat{x}}_0$, i.e., when $\mathcal F(\mathbf{\hat{x}}_0)=1$. It is often not possible to detect a face in $\mathbf{\hat{x}}_0$ with a large timestep $t$, at which $\mathcal F(\mathbf{\hat{x}}_0)=0$. The loss becomes:
\begin{equation}
\mathcal{L}=\mathcal{L}_{\mathrm{SD}}+\mathcal F(\mathbf 
{\hat{x}}_0) 
\cdot \lambda\mathcal{L}_\mathrm{id}
\end{equation}
where $\lambda$ controls for the weight of the face identity loss. Sec. \ref{sec::ablation} shows that face identity loss is effective in preserving output identity. We find that $\sim 50000$ iterations, $\lambda=0.1$ and learning rate $3\times10^{-5}$ is enough to train a robustly performing model.

%% file: 04_experiment.tex
\section{Experiments}
\label{sec:exp}

\subsection{Experimental settings}

\noindent\textbf{Datasets}\quad For our training process, we utilized the comprehensive collection of 30,000 image-text pairings from the Multi-Modal CelebA-HQ database, as detailed in \cite{xia2021tedigan}. This dataset includes 6,217 unique identities. To enhance the diversity of our dataset, we implemented various data augmentation techniques. These included random face swapping, utilizing the InsightFace~\cite{insightface} tool, alongside standard methods such as image flipping, rotation, and color adjustments. For each identity, we ensured the presence of over $N$ augmented images. During each iteration of training, $N$ images per identity were randomly chosen to generate the MFF vision embedding $\textbf E_r$ and the corresponding identity text embedding $\textbf E_c$.

For testing quantitative results, we methodically selected one image per individual for a total of 500 individuals from the VGGFace2 dataset~\cite{cao2018vggface2} as reference for all methods. For the measurement of identity preservation, our prompts for generation were limited to a simple ``\textit{[class noun]}" word such as "\textit{woman}" or "\textit{man}", and for the measurement of diversity, the prompts were a ``\textit{[class noun]}" word combined with a expression word (e.g. ``\textit{happy}", ``\textit{sad}", ``\textit{angry}"). It's noteworthy that all facial imagery used for visualization purposes were acquired from SFHQ dataset~\cite{SFHQ} or publicly accessible channels.

\vspace{\baselineskip}\noindent\textbf{Implementation Details}\quad We utilize Stable Diffusion~\cite{sd} V2.1 as the base model and fine-tune our IDAdapter at the training stage. We trained the model with Adam optimizer, learning rate of $3e-5$ and batch size of $4$ on a single A100 GPU. Our model was trained for $50,000$ steps. At the testing and inference stage, we use only one image and simply duplicate it $N$ times to serve as the input for the network.

\vspace{\baselineskip}\noindent\textbf{Evaluation Metrics}\quad A critical aspect in our evaluation is the fidelity of facial identity in the generated images. To quantify this, we calculate the average identity preservation, which is the pairwise cosine similarity between facial features of generated images and their real counterparts (ID-Sim). This calculation is performed using a pretrained face recognition model, as outlined in~\cite{deng2018arcface}. Additionally, we have introduced two novel metrics to assess the diversity of the generated images: pose-diversity (Pos-Div) and expression-diversity (Expr-Div).

\begin{itemize}

    \item \textbf{Pose-Diversity (Pose-Div)} This metric assesses the variance in facial angles between the generated image and the input image. To quantify this difference, we calculate the average deviation in facial angles across all test images. To better reflect real-world scenarios, we report the results specifically in terms of Pitch (Pose-Div pitch) and Yaw angles (Pose-Div yaw). This approach enables us to evaluate how well the model can generate images with a range of different facial orientations relative to the input image.
    \item \textbf{Expression-Diversity (Expr-Div)} This metric evaluates the variation in facial expressions between the generated images and the input image. Utilizing a pre-trained expression classification model, we measure the ratio of the generated images having different expression categories compared to the input across the entire test dataset. A higher value in this metric indicates a greater ability of the model to generate diverse facial expressions.
\end{itemize}
These metrics are crucial for determining the capability of our method to generate images that are not only personalized but also varied in terms of poses and expressions, reflecting a more comprehensive range of human facial appearance.

\subsection{Comparisons}

\begin{figure*}[tp]
    \centering
    \includegraphics[width=\textwidth]{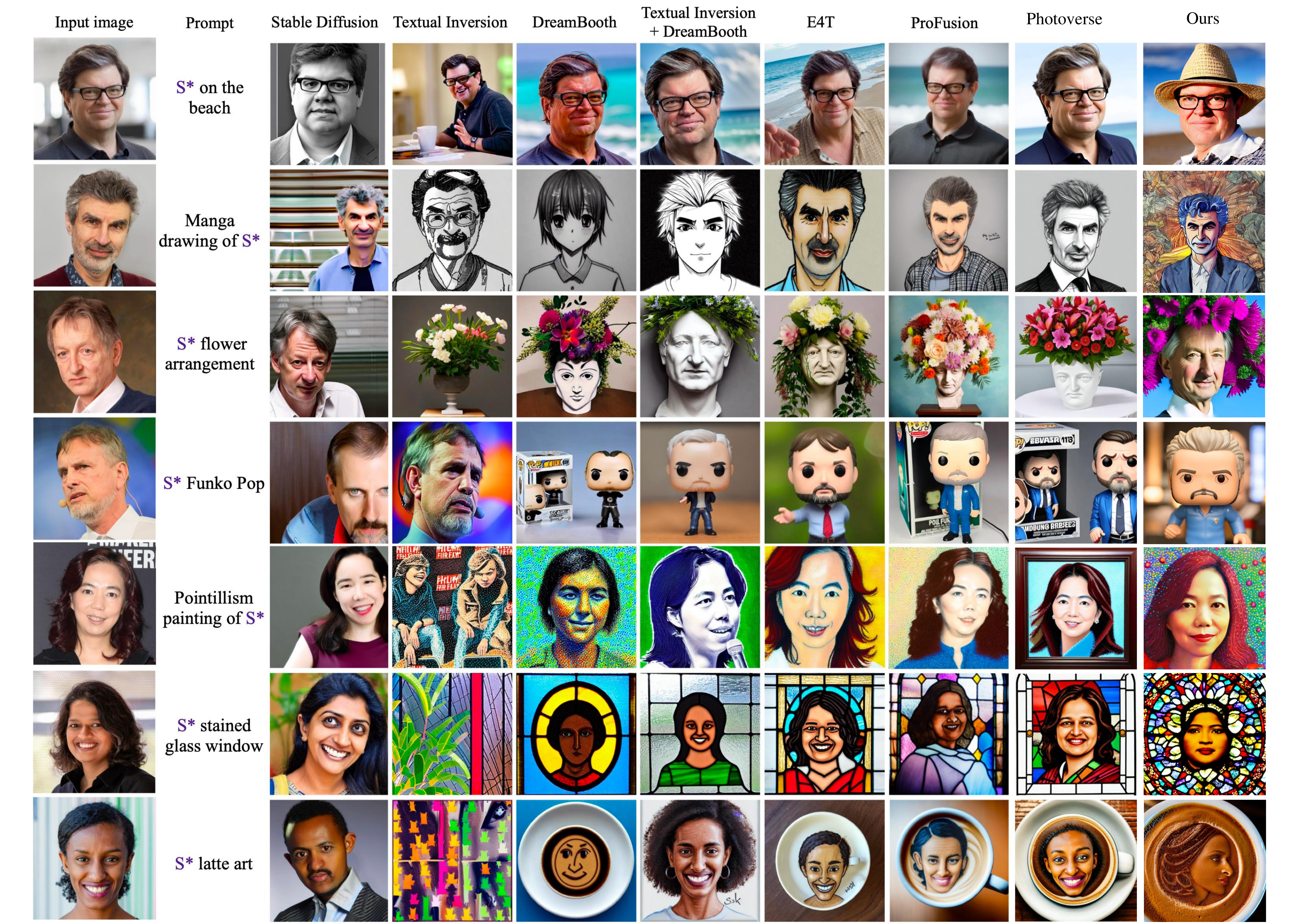}
    \caption{\textbf{Comparisons with several baseline methods.} IDAdapter is stronger in the diversity of properties, poses, expressions and other non-ID appearance, achieving very strong editability while preserving identity.}
    \label{fig::qua_comp}
\end{figure*}

\begin{figure*}[tp]
    \centering
    \includegraphics[width=\linewidth]{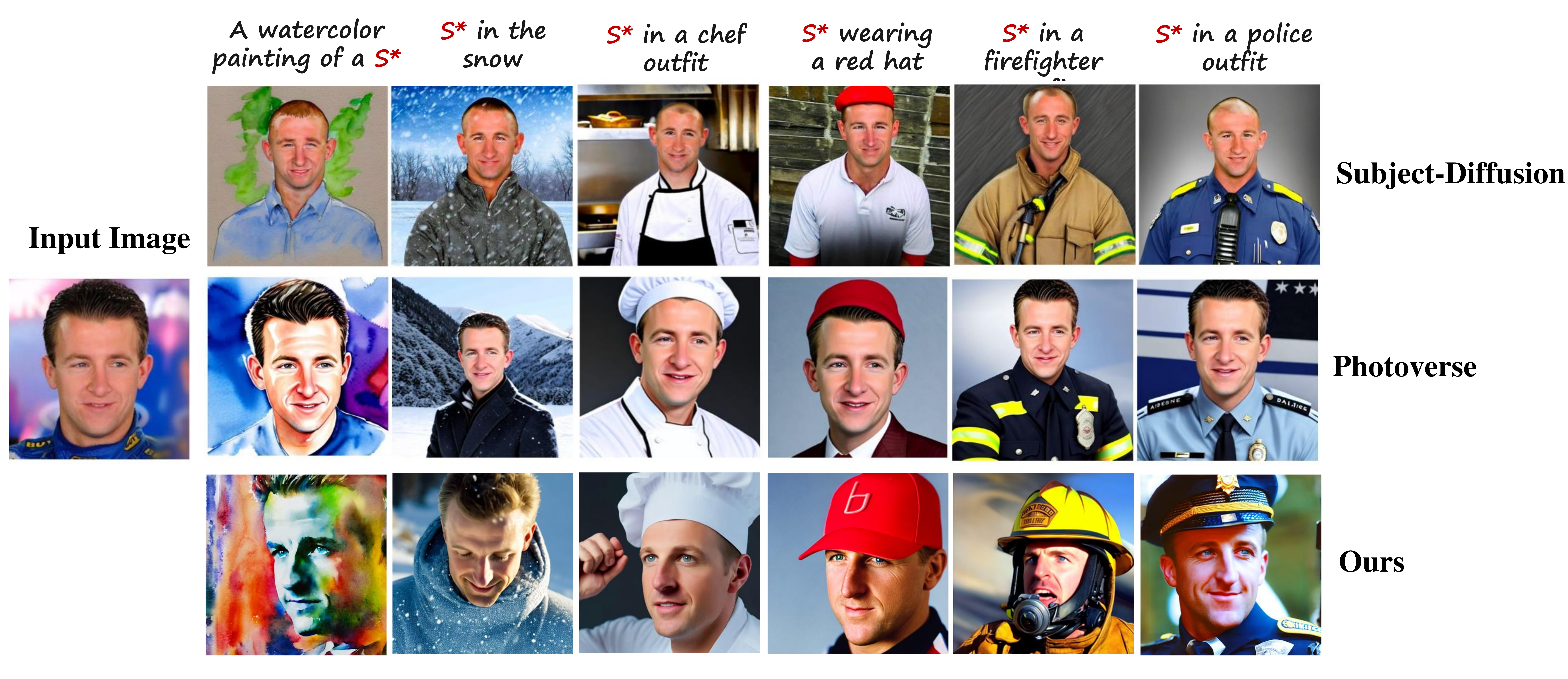}
    \caption{In terms of diversity performance, we compare generated samples of our method, Subject-Diffusion and Photoverse. We observe that our method generally achieves very strong diversity while preserving identity without giving in to overfitting to expressions or poses.}
    \label{fig::expr_comp}
\end{figure*}

\noindent\textbf{Qualitative Results}\quad Our methodology was benchmarked against several leading techniques, including Textual Inversion~\cite{gal2022image}, Dreambooth~\cite{ruiz2022dreambooth}, E4T~\cite{gal2023encoder}, ProFusion~\cite{zhou2023enhancing}, and Photoverse~\cite{chen2023photoverse}, as illustrated in Figure \ref{fig::qua_comp}. The comparative results were sourced directly from the study of ~\cite{chen2023photoverse}, where the ``\textit{S*}" in the prompts refers to the ``\textit{[class noun]}" we mentioned. We observe that our method surpasses both Textual Inversion and DreamBooth in terms of face fidelity. Unlike other methods, our approach effectively preserves identity without giving in to overfitting to expressions or poses as Figure \ref{fig::expr_comp} shows, thereby facilitating the generation of images that are both more diverse and lifelike.

\vspace{\baselineskip}\noindent\textbf{Quantitative Results}\quad In our quantitative experiments, the capability of IDAdapter was evaluated using three metrics: identity preservation (ID-Sim), pose-diversity (Pose-Div), and expression-diversity (Expr-Div). Moreover, these models demonstrate a lack of proficiency in generating varied facial expressions and poses. Consequently, we assessed Pos-Div and Expr-Div metrics exclusively on open-source models requiring fine-tuning \cite{gal2022image,ruiz2022dreambooth,kumari2023multi,yuan2023inserting}. In this experiment, we have selected the parameter $N=4$. As depicted in Table \ref{tab:quancomp}, our method achieved the highest scores across almost all metrics. It can be observed that IDAdapter effectively leverages the base model to generate more diverse results with identity preserved.

\begin{table*}[t]
\centering

\begin{tabular}{lcccccc}
 \toprule
\textbf{Method}   & \textbf{Fine-tuning} & \textbf{Single Image} & \textbf{ID-Sim ↑} & \textbf{Expr-Div ↑} & \textbf{Pose-Div pitch ↑} & \multicolumn{1}{c}{\textbf{Pose-Div yaw ↑}} \\ \midrule
Ours              & N                    & Y                     & 0.603           &  65\%                  & 7.90                      & 16.47                                        \\
Profusion \cite{gal2023encoder}       & Y                    & Y                     & 0.454           &     31\%              &         1.95                  &   2.31                                          \\
Celeb Basis \cite{yuan2023inserting}      & Y                    & Y                     & 0.207           &       35\%            &         4.92                  &   12.04                                          \\

DreamBooth  \cite{ruiz2022dreambooth}      & Y                    & N                     & 0.105           &       71\%            &           6.93                &   12.23                        \\ \bottomrule
\end{tabular}

\caption{
We compared our IDAdapter ($N=4$) with several baseline methods in terms of identity preservation (ID-Sim) and diversity performance (Expr-Div, Pose-Div pitch and Pose-Div yaw).
}
\vspace{-0.05in}
\label{tab:quancomp}
\end{table*}

\begin{figure*}[tp]
    \centering
    \includegraphics[width=\linewidth]{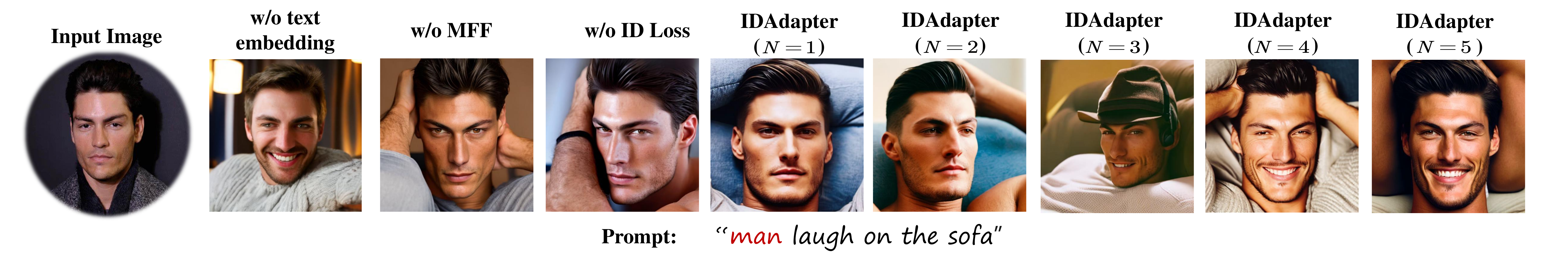}
    \caption{\textbf{Visualization of generated results under different settings.} Fine-tuning without certain model structure can result in a decrease in the performance of identity preservation and diversity, overfitting to input image appearance. MFF alleviates overfitting and help integrate detailed visual information into the model, allowing for more expression diversity and essential feature capture.}
    \label{fig::ablation}
\end{figure*}

\begin{table}[t]
\resizebox{\linewidth}{!}{
\begin{tabular}{lcccc}
  \toprule

\textbf{Method} & \textbf{ID-Sim ↑} & \textbf{Expr-Div ↑} & \makecell{\textbf{Pose-Div} \\ \textbf{pitch ↑}} & \makecell{\textbf{Pose-Div} \\ \textbf{yaw ↑}} \\ \midrule

No Text Embedding   & 0.394           &     49\%        &       6.08                    &         13.49                \\
No MFF          & 0.517           &      46\%        &       5.31                    &          13.26               \\
IDAdapter ($N=1$) & 0.602           &    37\%         &       5.02                    &         12.90                \\
IDAdapter ($N=2$) & 0.601           &    58\%          &     6.97                      &        15.39                 \\
IDAdapter ($N=3$) &     \textbf{0.604}       &    61\%         &      7.03                     &   15.44                      \\
IDAdapter ($N=4$) &0.603           &     \textbf{65\%}         & \textbf{7.90}                      & \textbf{16.47}                   \\
IDAdapter ($N=5$) & 0.601           &    64\%          &       7.88                    &          16.42               \\ \midrule
No ID Loss      & 0.592           &      57\%        &      7.64                     &           16.38              \\\bottomrule
\end{tabular}}
\caption{
Ablation studies on identity preservation metric (ID-Sim) and diversity metrics (Expr-Div, Pose-Div pitch and Pose-Div yaw).
}
\vspace{-0.05in}
\label{tab:ablation_mff}
\end{table}

\subsection{Ablation Studies}
\label{sec::ablation}
As illustrated in Table~\ref{tab:ablation_mff} and Figure~\ref{fig::ablation}, our analysis reveals the impact of different components of the IDAdapter method on the quality of generated images.

\noindent\textbf{Impact of Identity Text Embedding}\quad When the identity text embedding component is removed from the process (No Text Embedding), there is a significant decrease in the identity preservation of the generated images. This drastic drop suggests that textual conditions play a crucial role in guiding Stable Diffusion to generate personalized images. Without the identity text embedding, the fundamental feature of personalized generation is almost lost.

\noindent\textbf{Removal of MFF Vision Embedding}\quad Eliminating the vision embedding component output by MFF (No MFF) leads to a significant drop of both identity preservation and diversity. This indicates that the MFF module provides the model with rich identity-related content details. MFF is vital for counteracting overfitting and helps retain the ability of the base model to generate diverse images of the person.

\noindent\textbf{Impact of Different $N$ Values}\quad Changing the number of images $N$ used in training process has varying impacts on diversity and identity preservation. After testing with different $N$ values, we found that $N=4$ offers the best balance. It achieves a superior compromise between maintaining the identity similarity and enhancing the diversity. This balance is crucial for generating images that are both personalized and varied.

\noindent\textbf{Impact of ID Loss}\quad We trained IDAdapter ($N=4$) without face identity loss (No ID Loss). The model's performance in learning facial features has declined, and the generated faces are not as similar to the input as when incorporating the ID loss.

%% file: 10_conclusion.tex
\section{Conclusion}
\label{sec:conclusion}

We introduce a method named IDAdapter, which is the first to generate images of a person with a single input facial image in a variety of styles, angles, and expressions without fine-tuning during the inference stage, marking a significant breakthrough in personalized avatar generation.

%% file: 12_appendix.tex
\section{Implementation Details}
\vspace{\baselineskip}\noindent\textbf{Adapter Layer}\quad In our proposed approach, the adapter layer involves linear mapping of the MFF vision embedding, a gated self-attention mechanism, a feedforward neural network, and normalization before the attention mechanism and the feedforward network.

\vspace{\baselineskip}\noindent\textbf{Model Details}\quad 
The model in this work refers to the trainable structures, including the MFF module, a multi-layer perceptron, key and value projection matrices in each cross-attention block. The total model size is 262M, which is smaller than Subject Diffusion \cite{ma2023subjectdiffusionopen} (700M) and Dreambooth \cite{ruiz2022dreambooth} (983M). We set the sampling step as 50 for inference. Our method is tuning-free during testing, enabling the synthesis of 5 images within half a minute.

\section{Subject Personalization Results}
Our method achieve very effective editability, with semantic transformations of face identities into high different domains, and we conserve the strong style prior of the base model which allows for a wide variety of style generations. We show results in the following domains. The images for visualization is from SFHQ dataset \cite{SFHQ} and we use the unique facial image for each identity in the dataset as a reference to generate multiple images.

\begin{figure*}
    \centering
    \includegraphics[width=\linewidth]{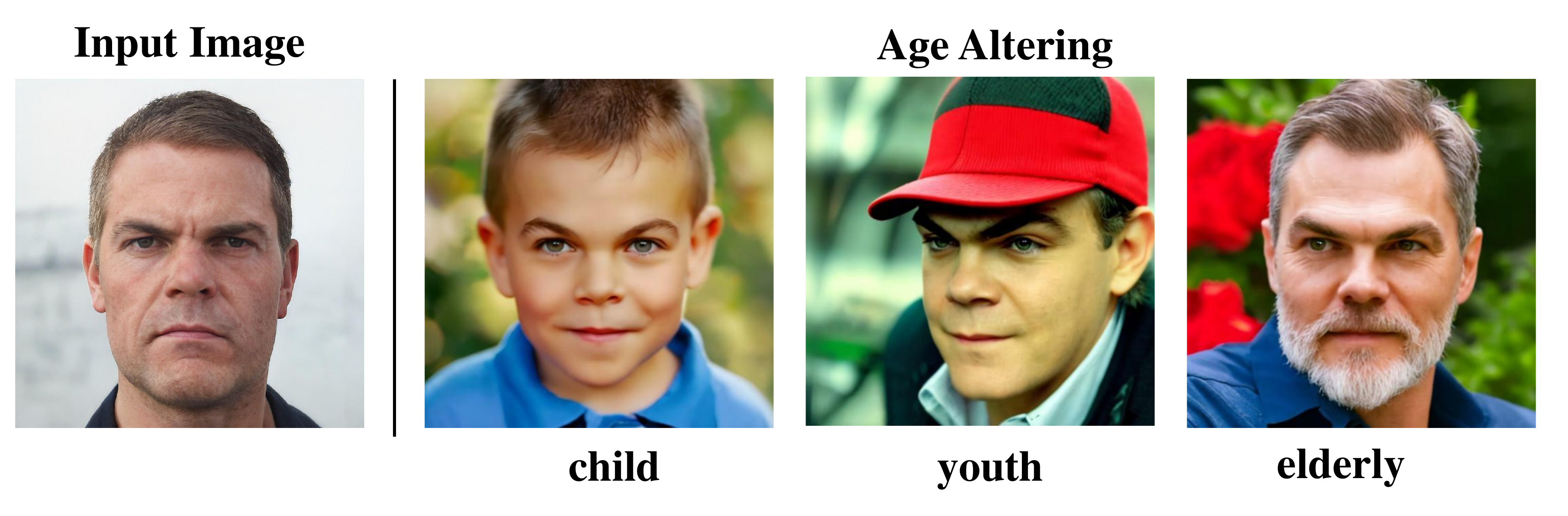}
    \caption{\textbf{Age altering.} We present photos of the same person at different age stages by prompting our generative model.}
    \label{fig::age}
\end{figure*}

\vspace{\baselineskip}\noindent\textbf{Age Altering}\quad We are able to generate novel faces of a person with different appearance of different age as Figure \ref{fig::age} shows, by including an age noun in the prompt sentence:``\textit{[class noun] is a [age noun]}". We can see in the example that the characteristics of the man is well preserved.

\begin{figure*}
    \centering
    \includegraphics[width=\linewidth]{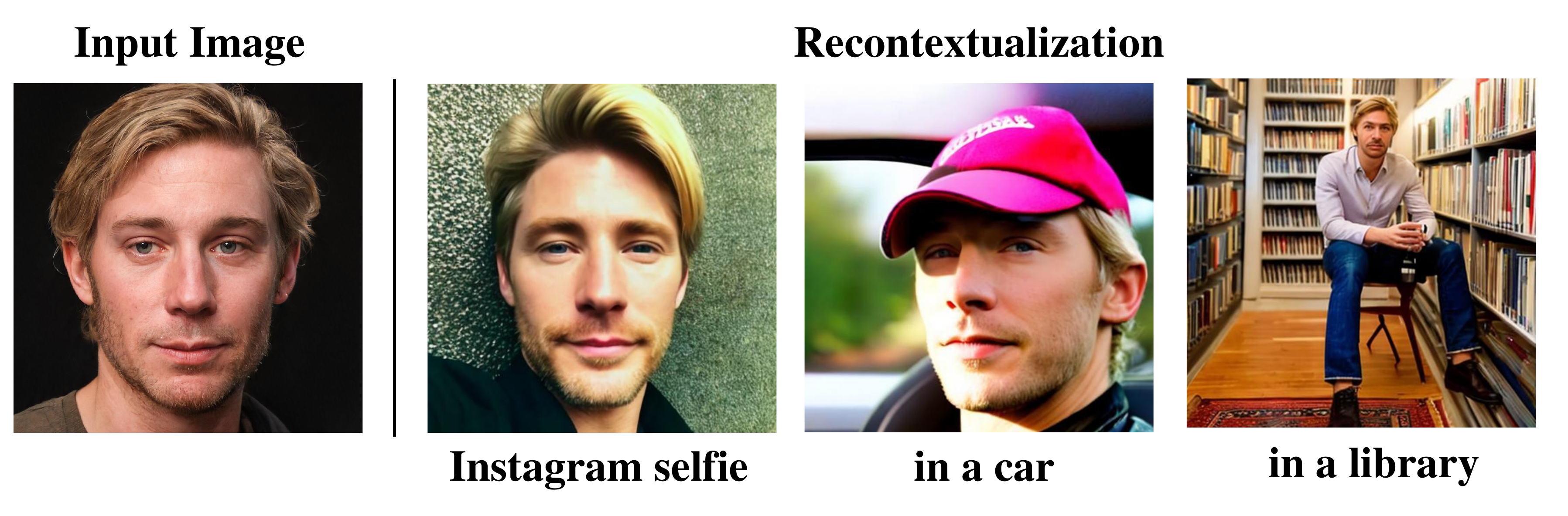}
    \caption{\textbf{Recontextualizaion.} We generate images of the subject person in different environments, with high preservation of facial details and realistic scene-subject interactions.}
    \label{fig::recontext}
\end{figure*}

\vspace{\baselineskip}\noindent\textbf{Recontextualization}\quad We can generate novel images for a specific person in different contexts (Figure \ref{fig::recontext}) with descriptive prompts (``\textit{a [class noun] [context description]}"). Importantly, we are able to generate the person in new expressions and poses, with previously unseen scene structure and realistic integration of the person in the scene.

\begin{figure*}
    \centering
    \includegraphics[width=\linewidth]{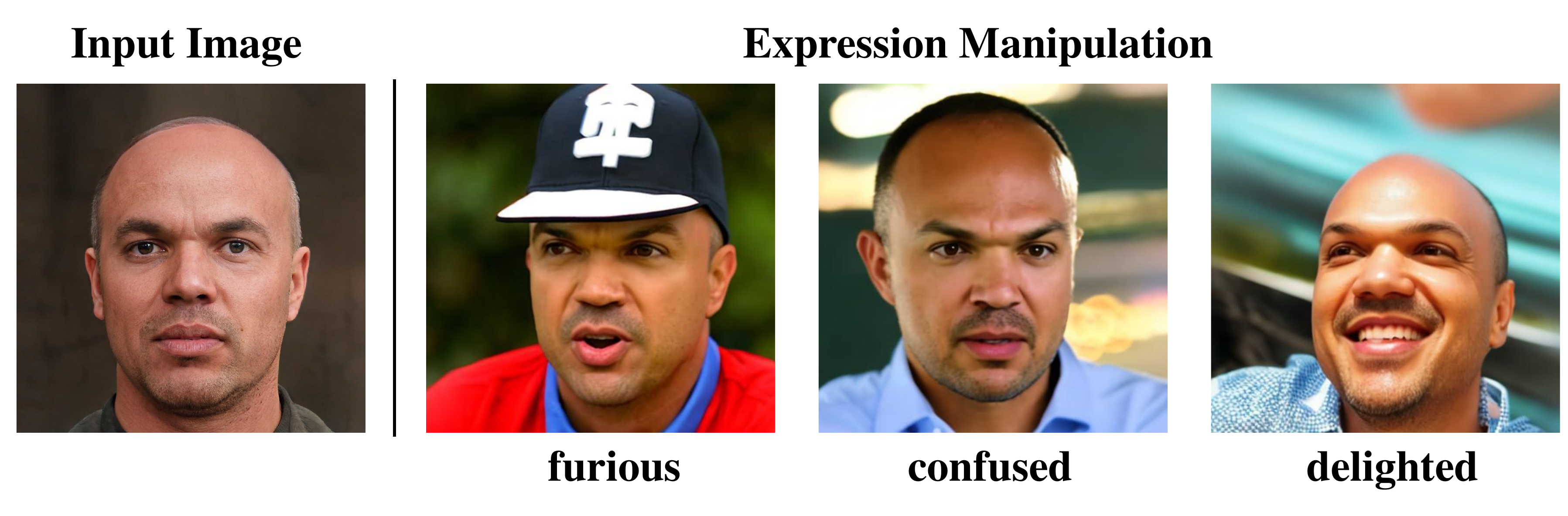}
    \caption{\textbf{Expression manipulation.} Our method can generate a range of expressions not present in the input images, showcasing the model's inference capabilities.}
    \label{fig::expr}
\end{figure*}

\vspace{\baselineskip}\noindent\textbf{Expression Manipulation}\quad Our method allows for new image generation of a person with modified expressions that are not seen in the original input image by prompts ``\textit{[class noun] is [expression adjective]}". We show examples in Figure \ref{fig::expr}. 

\begin{figure*}
    \centering
    \includegraphics[width=\linewidth]{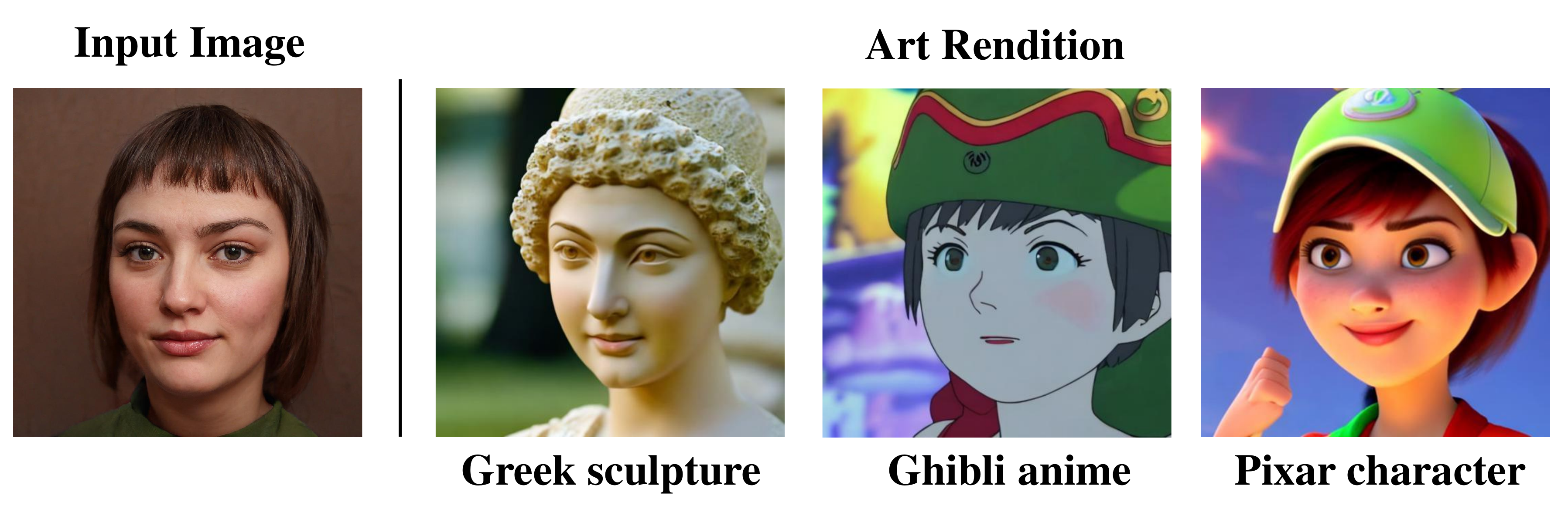}
    \caption{\textbf{Artistic renderings.} We can observe significant changes in the appearance of the character to blend facial features with the target artistic style. }
    \label{fig::art}
\end{figure*}

\vspace{\baselineskip}\noindent\textbf{Art Renditions}\quad Given a prompt ``\textit{[art form] of [class noun]}", we are able to generate artistic renditions of the person. We show examples in Figure \ref{fig::art}. We select similar viewpoints for effect, but we can generate different angles of the woman with different expressions actually.

\begin{figure*}
    \centering
    \includegraphics[width=\linewidth]{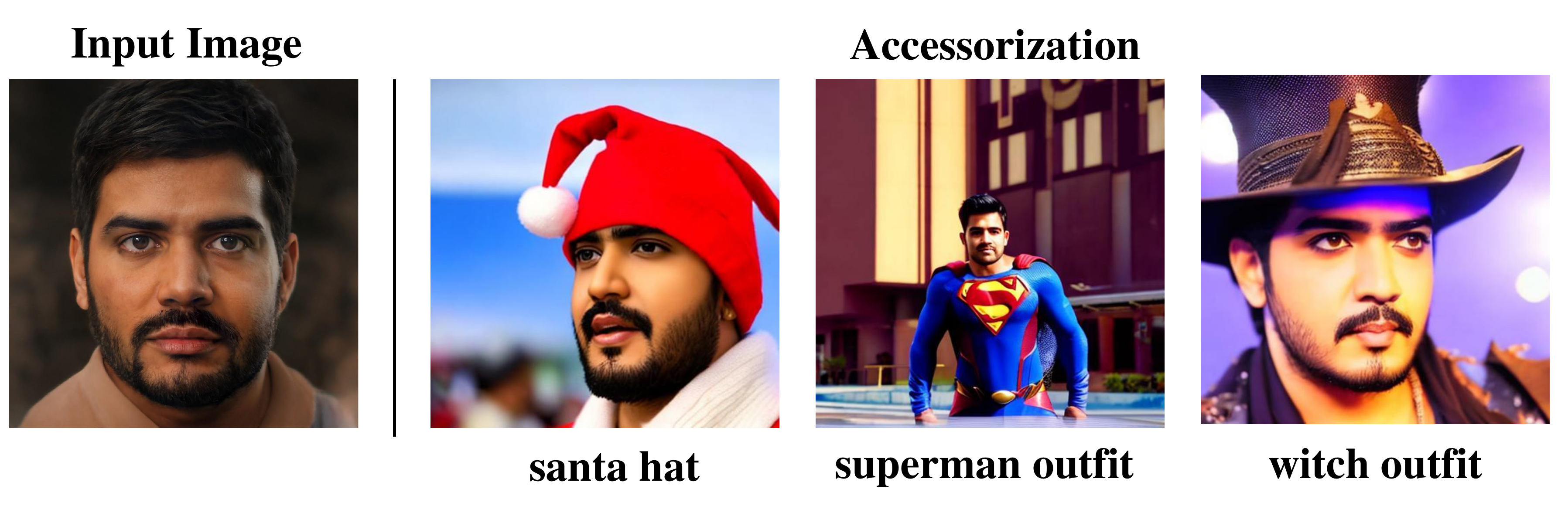}
    \caption{\textbf{Outfitting a man with accessories.} The identity of the subject person is preserved and different outfits or accessories can be applied to the man given a prompt of type ``\textit{[class noun] in [accessory]}". We observe a realistic interaction between the subject man and the outfits or accessories. }
    \label{fig::access}
\end{figure*}

\vspace{\baselineskip}\noindent\textbf{Accessorization}\quad We utilize the capability of the base model to accessorize subject persons. In Figure \ref{fig::access}, we show examples of accessorization of a man. We prompt the model with a sentence:``\textit{[class noun] in [accessory]}" to fit different accessories onto the man with aesthetically pleasing results.

\begin{figure*}
    \centering
    \includegraphics[width=\linewidth]{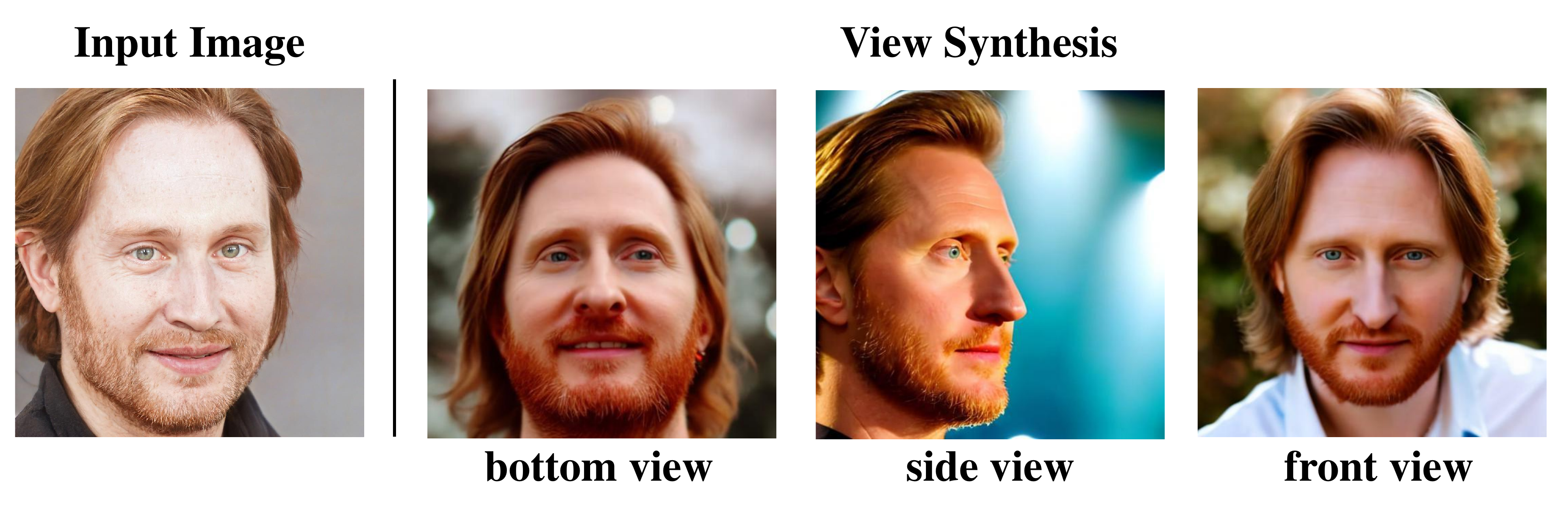}
    \caption{\textbf{View Synthesis.} Our technique can synthesize images with specified viewpoints for a subject person.}
    \label{fig::view}
\end{figure*}

\vspace{\baselineskip}\noindent\textbf{View Synthesis}\quad We show several viewpoints for facial view synthesis in Figure \ref{fig::view}, using prompts as "\textit{[class noun] [viewpoint]}" in the figure.

\begin{figure*}
    \centering
    \includegraphics[width=\linewidth]{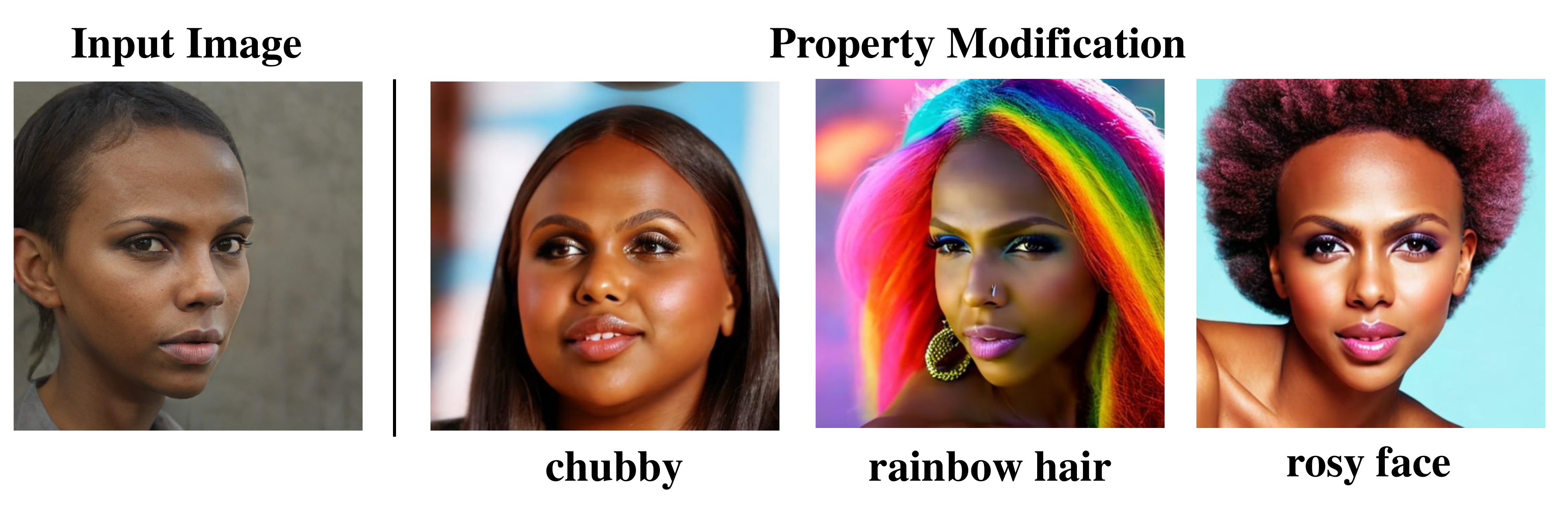}
    \caption{\textbf{Modification of subject properties.} We show modifications in the body type, hair color and complexion (using prompts ``\textit{[class noun] is/has [property description]}").}
    \label{fig::property}
\end{figure*}

\vspace{\baselineskip}\noindent\textbf{Property Modification}\quad We are able to modify facial properties. For example, we show a different body type, hair color and complexion in Figure \ref{fig::property}. We prompt the model with the sentences ``\textit{[class noun] is/has [property description]}". In particular, we can see that the identity of the face is well preserved.

\begin{figure*}
    \centering
    \includegraphics[width=\linewidth]{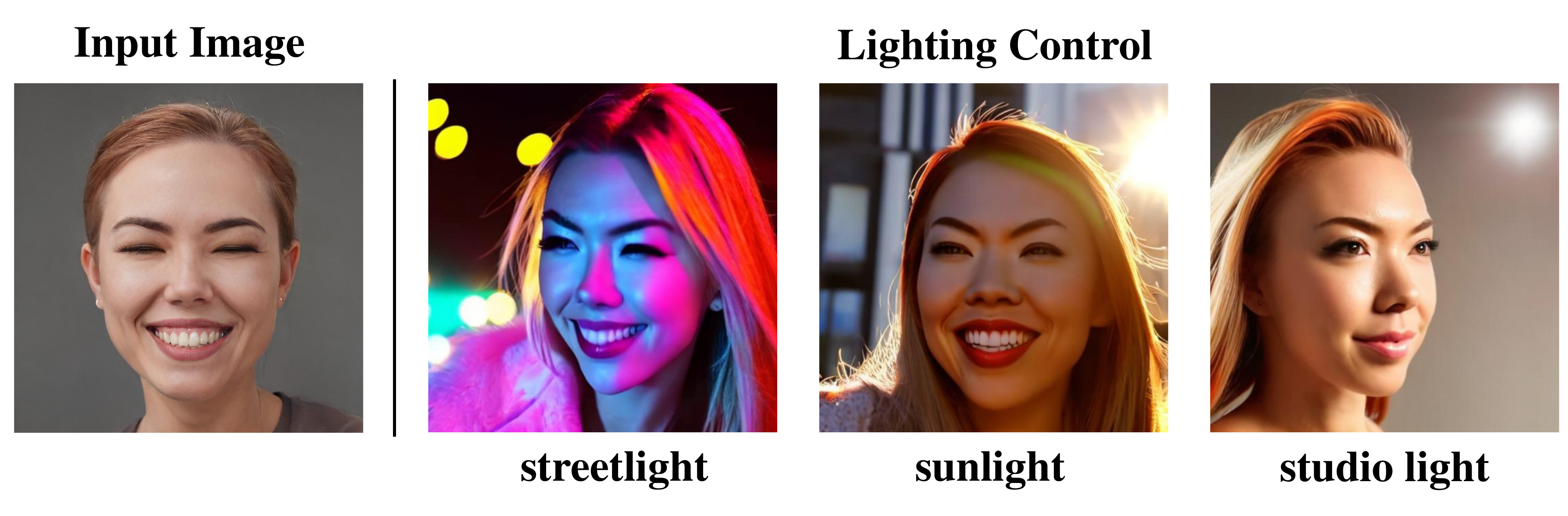}
    \caption{\textbf{Lighting control.} Our method can generate lifelike subject photos under different lighting conditions, while maintaining the integrity to the subject’s key facial characteristics.}
    \label{fig::light}
\end{figure*}

\vspace{\baselineskip}\noindent\textbf{Lighting control}\quad Our personalization results exhibit natural variation in lighting and we can also control the lighting condition by prompts like ``\textit{[class noun] in [lighting condition]}", which may not appear in the reference images. We show examples in Figure \ref{fig::light}.

\begin{figure*}
    \centering
    \includegraphics[width=\linewidth]{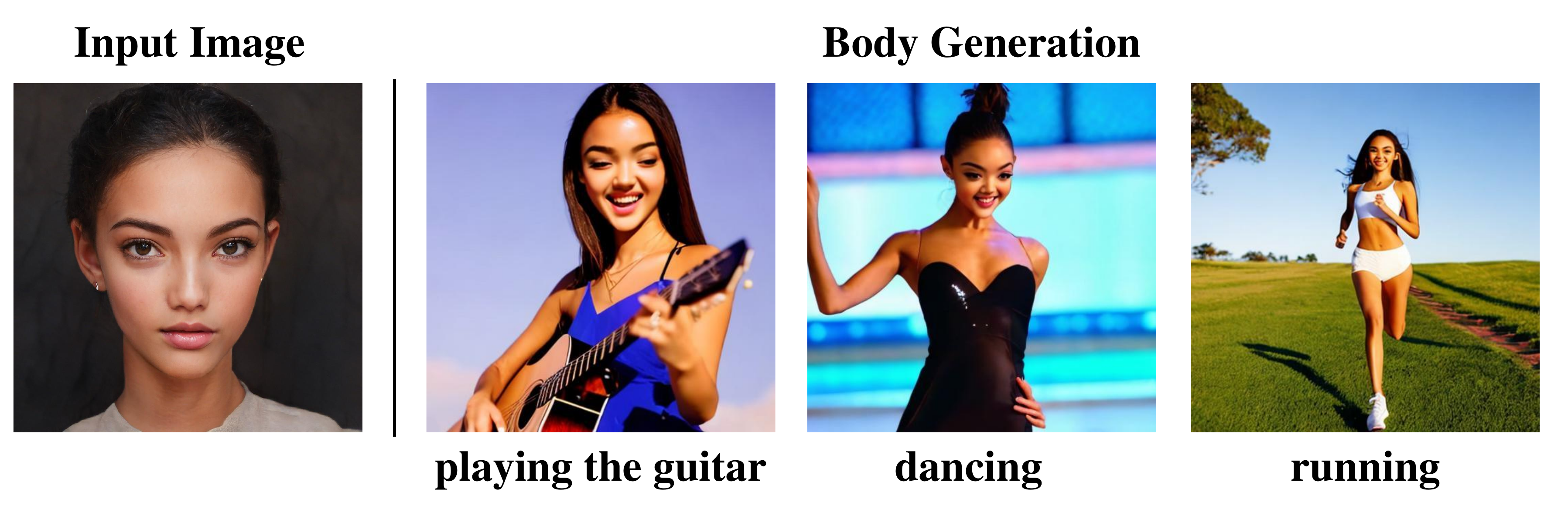}
    \caption{\textbf{Body generation.} We are able to generate the body of the subject person in novel poses and articulations with only a facial image.}
    \label{fig::body}
\end{figure*}

\vspace{\baselineskip}\noindent\textbf{Body Generation}\quad Our model has the ability to infer the body of the subject person from facial features and can generate specific poses and articulations in different contexts based on the prompts combined with ``\textit{full/upper body shot}" as Figure \ref{fig::body} shows. In essence, we seek to leverage the model's prior of the human class and entangle it with the embedding of the unique identifier.

